\documentclass{article}

\PassOptionsToPackage{numbers, compress}{natbib}

\usepackage[preprint]{neurips_2023}

\usepackage[utf8]{inputenc} 
\usepackage[T1]{fontenc}    
\usepackage{hyperref}       
\usepackage{url}            
\usepackage{booktabs}       
\usepackage{amsfonts}       
\usepackage{nicefrac}       
\usepackage{microtype}      
\usepackage{xcolor}         
\usepackage{graphicx}%
\usepackage{amsmath,amssymb}%
\usepackage{bbding} 
\usepackage{multirow}%
\usepackage{amsmath,amssymb,amsfonts}%
\usepackage{amsthm}%
\usepackage{mathrsfs}%
\usepackage[title]{appendix}%
\usepackage{xcolor}%
\usepackage{textcomp}%
\usepackage{booktabs}%
\usepackage{algorithm}%
\usepackage{algorithmicx}%
\usepackage{algpseudocode}%
\usepackage{listings}%
\usepackage{subfig}
\usepackage{balance}
\usepackage[hang,flushmargin]{footmisc}

\makeatletter
\def\blfootnote{\xdef\@thefnmark{}\@footnotetext}
\makeatother

\title{Matten: Video Generation with Mamba-Attention}

\author{
	\hspace{-0.25cm}\textbf{Yu Gao$^{1}$, Jiancheng Huang$^{1}$, Xiaopeng Sun$^{1}$, Zequn Jie$^{\dagger 1}$}, Yujie Zhong$^{1}$ \\
	\hspace{-0.25cm}\textbf{Lin Ma$^{1}$} \\ 
	\hspace{-0.25cm}$^1$Meituan Inc. 
}

\begin{document}

\maketitle

\thispagestyle{empty}

 \blfootnote{\noindent
$^{\dagger}$ Corresponding to Zequn Jie <zequn.nus@gmail.com>.}

\begin{abstract}
In this paper, we introduce Matten, a cutting-edge latent diffusion model with Mamba-Attention architecture for video generation. With minimal computational cost, Matten employs spatial-temporal attention for local video content modeling and bidirectional Mamba for global video content modeling. Our comprehensive experimental evaluation demonstrates that Matten has competitive performance with the current Transformer-based and GAN-based models in benchmark performance, achieving superior FVD scores and efficiency. Additionally, we observe a direct positive correlation between the complexity of our designed model and the improvement in video quality, indicating the excellent scalability of Matten.
\end{abstract}

\section{Introduction}
\begin{figure*}[t]  
\centering
\includegraphics[width=\linewidth]{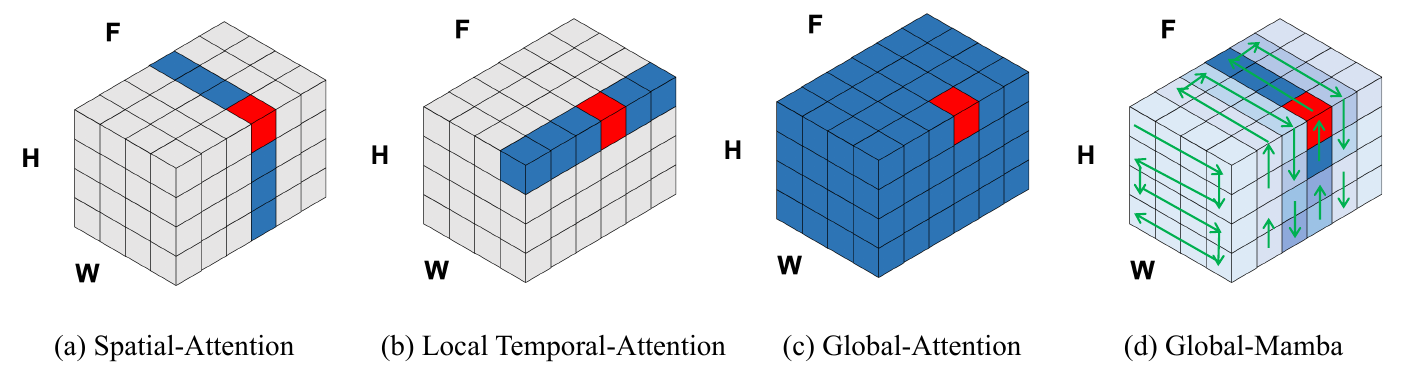}
\caption{Different ways of spatio-temporal modeling using Mamba and Attention. $H$, $W$, and $F$ denote the height, weight, and frames, respectively. The red token is an example query, and the blue tokens mean those tokens having information interaction with the query. The shade of blue represents the intensity of the information interaction, with darker colors representing more direct interactions. Mamba scan interactions are distance-related between tokens with a linear complexity, while attention interactions are equal among these tokens with a quadratic complexity. For simplicity, we only show the unidirectional Mamba scan.}
\label{fig_homepage}
\end{figure*}

Recent advancements in diffusion models have demonstrated impressive capabilities in video generation \cite{yu2023video, ma2024latte, lu2023vdt, ho2022video, voleti2022mcvd}. It has been observed that breakthroughs in architectural design are crucial for the efficient application of these models \cite{nichol2021improved, peebles2023scalable, bao2023all}. Contemporary studies largely concentrate on CNN-based U-Net architectures \cite{yu2023video, ho2022video} and Transformer-based frameworks \cite{lu2023vdt, ma2024latte}, both of which employ attention mechanisms to process spatio-temporal dynamics in video content. Spatial attention, which involves computing self-attention among image tokens within a single frame, is extensively utilized in both U-Net-based and Transformer-based video generation diffusion models as shown in Fig. \ref{fig_homepage} (a). Prevailing techniques typically apply local attention within the temporal layers as illustrated in Fig. \ref{fig_homepage} (b), where attention calculations are confined to identical positions across different frames. This approach fails to address the critical aspect of capturing interrelations across varying spatial positions in successive frames. A more effective method for temporal-spatial analysis would involve mapping interactions across disparate spatial and temporal locations, as depicted in Fig. \ref{fig_homepage} (c). Nonetheless, this global-attention method is computationally intensive due to the quadratic complexity involved in computing attention, thus requiring substantial computational resources.

There has been a rise in fascination with state space models (SSMs) across a variety of fields, largely due to their ability to deal with long sequences of data \cite{smith2022simplified, gu2023mamba, gu2021combining}. In the field of Natural Language Processing (NLP), innovations such as the Mamba model \cite{gu2023mamba} have significantly improved both the efficiency of data inference processes and the overall performance of models by introducing dynamic parameters into the SSM structure and by building algorithms tailored for better hardware compatibility. The utility of the Mamba framework has been successfully extended beyond its initial applications, demonstrating its effectiveness in areas such as vision \cite{liu2024vmamba, zhu2024vision} and multimodal applications \cite{zhao2024cobra}. Given the complexity of processing video data, we propose to use the Mamba architecture to explore spatio-temporal interactions in video content, as shown in Fig. \ref{fig_homepage} (d). However, unlike the self-attention layer, it's important to note that Mamba scans, which do not inherently compute dependencies between tokens, struggle to effectively detect localised data patterns, a limitation pointed out by \cite{zuo2022efficient}.

Regarding the advantages of Mamba and Attention, we introduce a latent diffusion model for video generation with a Mamba-Attention architecture, namely \textbf{Matten}. Specifically, we investigated the impact of various combinations of Mamba and Attention mechanisms on video generation. Our findings demonstrate that the most effective approach is to utilize the Mamba module to capture global temporal relationships (Fig. \ref{fig_homepage} (d)) while employing the Attention module for capturing spatial and local temporal relationships (Fig. \ref{fig_homepage} (a) and Fig. \ref{fig_homepage} (b)).

We conducted experimental evaluations to examine the performance and effects of Matten in both unconditional and conditional video generation tasks. Across all test benchmarks, Matten consistently exhibits the comparable FVD score \cite{unterthiner2018towards} and efficiency with SOTAs. Furthermore, our results indicate that Matten is scalable, evidenced by the direct positive relationship between the model's complexity and the quality of generated samples. 

In summary, our contributions are as follows: 
\setlength{\itemsep}{0pt}
\setlength{\parsep}{0pt}
\setlength{\parskip}{0pt}
\begin{itemize}
\item  We propose Matten, a novel video latent diffusion model integrated with the mamba block and attention operations, which enables efficient and superior video generation. 

\item We design four model variants to explore the optimal combination of Mamba and attention in video generation. Based on these variants, we find that the most favorable approach is adopting attention mechanisms to capture local spatio-temporal details and utilizing the Mamba module to capture global information. \item  Comprehensive evaluations show that our Matten achieves comparable performance to other models with lower computational and parameter requirements and exhibits strong scalability.
\end{itemize}

\section{Related Work}

\subsection{Video Generation}
The task of video generation primarily focuses on producing realistic video clips characterized by high-quality visuals and fluid movements. Previous video generation work can be grouped into 3 types~\cite{melnik2024video,xing2023survey}. Initially, a number of researchers focused on adapting powerful GAN-based image generation techniques for video creation~\cite{vondrick2016generating, saito2017temporal, wang2020imaginator, wang2020g3an, kahembwe2020lower}. Nonetheless, GAN-based methods may lead to problems such as mode collapse, reducing diversity and realism.

In addition, certain models suggest the learning of data distributions via autoregressive models~\cite{ge2022long, rakhimov2020latent, weissenborn2019scaling, yan2021videogpt}. These methods typically yield high-quality videos and demonstrate more reliable convergence, but they are hindered by their substantial computational demands.
Finally, the latest strides in video generation are centered on the development of systems that utilize diffusion models~\cite{ho2020denoising, harvey2022flexible, ho2022video, singer2022make, mei2023vidm, blattmann2023align, wang2023lavie, chen2023seine, wang2023leo, ma2024latte}, which have shown considerable promise. These methods primarily use CNN-based U-Net or Transformer as the model architecture.
Distinct from these works, our method concentrates on investigating the underexplored area of the combination of mamba and attention within video diffusion.

\subsection{Mamba}
Mamba, a new State-Space Model, has recently gained prominence in deep learning~\cite{wang2024state,xu2024survey,zhang2024survey,patro2024mamba,liu2024vision} for its universal approximation capabilities and efficient modeling of long sequences, with applications in diverse fields such as medical imaging, image restoration, graphs, NLP, and image generation~\cite{ma2024u, wang2024mambabyte, zheng2024u, liang2024pointmamba, fei2024scalable, behrouz2024graph, ahamed2024mambatab}. Drawing from control systems and leveraging HiPPO initialization~\cite{gu2020hippo}, these models, like LSSL~\cite{gu2021combining}, address long-range dependencies but are limited by computational demands. To overcome this, S4 ~\cite{gu2021efficiently} and other structured state-space models introduce various configurations~\cite{gupta2022diagonal, gu2022parameterization, smith2022simplified} and mechanisms~\cite{gu2023mamba} that have been integrated into larger representation models~\cite{mehta2022long, ma2022mega, fu2022hungry} for tasks in language and speech. Mamba, and its iterations like VisionMamba~\cite{liu2024vmamba, zhu2024vision}, S4ND~\cite{nguyen2022s4nd}, and Mamba-ND~\cite{li2024mamba}, exhibit a range of computational strategies, from bidirectional SSMs to local convolution and multi-dimensionality considerations. For 3D imaging, T-Mamba~\cite{hao2024t} tackles the challenges in orthodontic diagnosis with Mmaba to handle long-range dependencies. For video understanding, VideoMamba~\cite{li2024videomamba} and Video Mamba Suite~\cite{chen2024video} adapt Mamba to the video domain and address the challenges of local redundancy and global dependencies prevalent in video data. 
In the domain of diffusion applications using mamba, Zigzag Mamba~\cite{hu2024zigma} advances the scalability and efficiency of generating visual content. It tackles the crucial problem of spatial continuity with an innovative scanning approach, incorporates text-conditioning features, and shows enhanced performance across high-resolution image and video datasets.
~\cite{oshima2024ssm} closely relates to our work, employing the mamba block in the temporal layer of video diffusion. Diverging from previous research focused mainly on local temporal modeling, our method, Matten, is uniquely designed to encompass global temporal dimensions. 

\section{Methodology}
\label{methodology}
Our discussion starts with a brief overview of the latent space diffusion model and state space model in Sec. \ref{sec:preliminary_of_diffusion_models}. This is followed by an in-depth description of the Matten model variants in Sec. \ref{sec:the_details_of_Matten}. We then explore conditional ways related to timestep or class in Sec. \ref{sec:conditional_information_injection}. Lastly, a theoretical analysis comparing Mamba with Attention mechanisms is presented in Sec. \ref{sec:compuation_anaylysis}.

\subsection{Background}

\label{sec:preliminary_of_diffusion_models}

\textbf{Latent Space Diffusion Models.} \cite{rombach2022high}.
For an input data sample \( x \in p_{\text{data}}(x) \), Latent Diffusion Models (LDMs) initially utilize the pre-trained VAE or VQ-VAE encoder \( \mathcal{E} \) to transform the data sample into a latent representation \( z = \mathcal{E}(x) \). This transformation is followed by a learning phase where the data distribution is modeled through diffusion and denoising steps.

During the diffusion phase, noise is incrementally added to the latent encoding, producing a series of increasingly perturbed latent states \( z_{t} \), where the intensity of additive noise is denoted by the timesteps \( t \in T \). A specialized model such as U-Net \( \epsilon_{\theta} \) is utilized as the noise estimate network to estimate the noise perturbations affecting the latent representation \( z_{t} \) during the denoising phase, aiming to minimize the latent diffusion objective.
\begin{equation}
\mathcal{L}_{simple} = \mathbb{E}_{\mathbf{z}\sim p(z),\ \epsilon \sim \mathcal{N} (0,\mathbf{I}),\ t}\left [ \left \| \epsilon - \epsilon_{\theta}(\mathbf{z}_t, t)\right \|^{2}_{2}\right].
\end{equation}

Furthermore, the diffusion models \( \epsilon_{\theta} \) are enhanced with a learned reverse process covariance \( \Sigma_\theta \), optimized using \( \mathcal{L}_{vlb} \) as outlined by \cite{nichol2021improved}.
\begin{figure*}[!t]
    \centering
    \includegraphics[width=\linewidth]{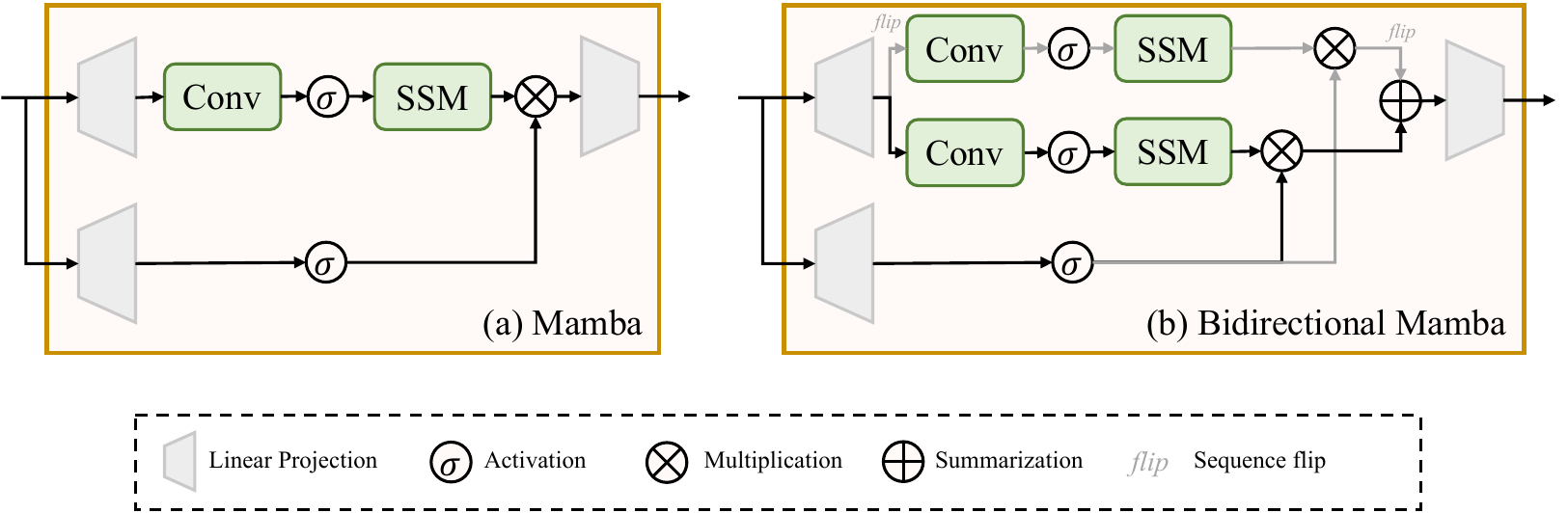}
    \caption{The original 1D sequence Mamba block and 2D bidirectional Mamba block. The normalization and the residual are omitted for simplification.} 
    \label{fig_mamba}
\end{figure*}

In our research, \( \epsilon_{\theta} \) is designed using a Mamba-based framework. Both \( \mathcal{L}_{simple} \) and \( \mathcal{L}_{vlb} \) are employed to refine the model's effectiveness and efficiency.

\textbf{State Space Backbone.}\label{sec:ssm} 
State space models (SSMs) have been rigorously validated both theoretically and through empirical evidence to adeptly manage long-range dependencies, demonstrating linear scaling with the length of data sequences. Conventionally, a linear state space model is represented as the following type:
\begin{equation}
\begin{aligned}
    h'(t) &= \mathbf{A}(t)h(t) + \mathbf{B}(t)x(t), \\
    y(t) &= \mathbf{C}(t)h(t) + \mathbf{D}(t)x(t),
\end{aligned}
    \label{eq:ssm-con}
\end{equation}
which describes the transformation of a 1-D input sequence \(x(t) \in \mathbb{R}\) into a 1-D output sequence \(y(t) \in \mathbb{R}\), mediated by an N-D latent state sequence \(h(t) \in \mathbb{R}^N\). State space models are particularly crafted to integrate multiple layers of these basic equations within a neural sequence modeling architecture, allowing the parameters \(\mathbf{A}, \mathbf{B}, \mathbf{C}\), and \(\mathbf{D}\) of each layer to be optimized via deep learning on loss function. $N$ represents the state size, ${\rm \mathbf{A}} \in \mathbb{R}^{N\times N}$, ${\rm \mathbf{B}} \in \mathbb{R}^{N \times 1}$, ${\rm \mathbf{C}} \in \mathbb{R}^{1\times N}$, and ${\rm \mathbf{D}} \in \mathbb{R}$.

The process of discretization, essential for applying state space models as detailed in Eq.~\ref{eq:ssm-con} to real-world deep learning tasks, converts continuous system parameters like \(\mathbf{A}\) and \(\mathbf{B}\) into their discrete equivalents \(\overline{\mathbf{A}}\) and \(\overline{\mathbf{B}}\). This critical step typically utilizes the zero-order hold (ZOH) method, a technique well-established in academic research for its efficacy. The ZOH method uses the timescale parameter \(\Delta\) to bridge the gap between continuous and discrete parameters, thereby facilitating the application of theoretical models within computational settings.

\begin{equation}
\begin{aligned}
    \overline{\mathbf{A}} &= \exp(\Delta \mathbf{A}),\\
    \overline{\mathbf{B}} &= (\Delta \mathbf{A})^{-1}(\exp(\mathbf{A}) - \mathbf{I}) \cdot \Delta \mathbf{B}.
\end{aligned}
\end{equation}

With these discretized parameters, the model outlined in Eq.~\ref{eq:ssm-con} is then adapted to a discrete framework using a timestep \(\Delta\):

\begin{equation}
\begin{aligned}
\label{eq:discret-ssm}
    h_k &= \overline{\mathbf{A}} h_{k-1} + \overline{\mathbf{B}} x_k,\\
    y_k &= \mathbf{C} h_k + \mathbf{D} x_k.
\end{aligned}
\end{equation}

This approach allows for the seamless integration of state space models into digital platforms. The traditional Mamba block, initially crafted for 1D sequence processing as shown in Fig.~\ref{fig_mamba}, is not ideally suited for visual tasks that demand spatial cognizance. To address this limitation, Vision Mamba~\cite{zhu2024vision} has developed a bidirectional Mamba block specifically tailored for vision-related applications. This innovative block is engineered to handle flattened visual sequences by employing both forward and backward SSMs concurrently, significantly improving its ability to process with spatial awareness.

Mamba employs a work-efficient parallel scan that effectively reduces the sequential dependencies typically associated with recurrent computations. This optimization, coupled with the strategic utilization of GPU operations, eliminates the necessity to explicitly manage the expanded state matrix. In our study, we explore the integration of the Mamba architecture within a video generation framework, leveraging its efficiency and scalability.

\begin{figure*}[!t]
    \centering
    \includegraphics[width=\linewidth]{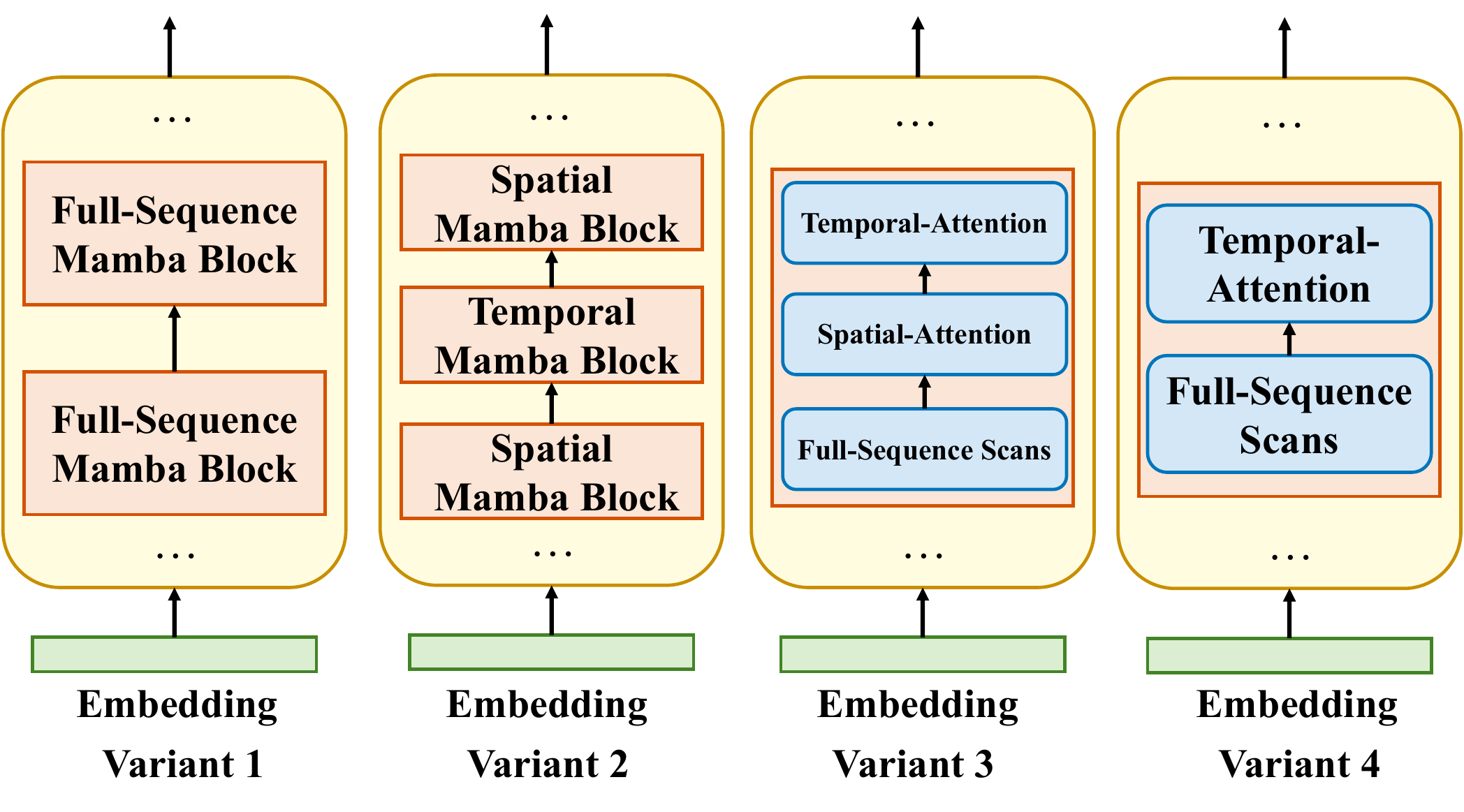}
    \caption{We introduce four model variants designed to harness spatio-temporal dynamics in videos effectively. For clarity, the embeddings shown in the diagram represent the patch and reshaped outcomes of the latent video.} 
    \label{fig_architecture}
\end{figure*}
\subsection{The Model Variants of Matten}
\label{sec:the_details_of_Matten}
Consider the representation of a video clip's latent space, represented by \(\boldsymbol{V_L} \in \mathbb{R}^{F \times H \times W \times C}\), where \(F\) indicates the number of frames, \(H\) the height of the frame, \(W\) the width of the frame, and \(C\) the channels per frame within the video's latent configuration. We transform \(\boldsymbol{V_L}\) into a sequence of tokens by segmenting and reshaping it, represented as \(\hat{\boldsymbol{z}} \in \mathbb{R}^{(n_f \times n_h \times n_w) \times d}\). Here, \(n_f \times n_h \times n_w\) denotes the total number of tokens, with each token having dimension \(d\).

Adopting a strategy similar to Latte, we assign \(n_f = F\), \(n_h = H/2\), and \(n_w = W/2\) to structure the data effectively. Furthermore, a spatio-temporal positional embedding, denoted as \(\boldsymbol{p}\), is incorporated into the token sequence \(\hat{\boldsymbol{z}}\). The input for the Matten model thus becomes \(\boldsymbol{z} = \hat{\boldsymbol{z}} + \boldsymbol{p}\), facilitating complex model interactions. As illustrated in Fig. \ref{fig_architecture}, we introduce four distinct variants of the Matten model to enhance its versatility and effectiveness in video processing.

\textbf{Global-Sequence Mamba Block.}
As illustrated in Fig. \ref{fig_architecture} (a), this variant refers to the execution of 3D Mamba scans in the full sequence of this spatiotemporal input. Following VideoMamba~\cite{li2024videomamba}, we adopt \textit{Spatial-First Scan} for our Global-Sequence Mamba block. 
This straightforward operation has been proven to be highly effective. It involves arranging spatial tokens based on their location and stacking them sequentially frame by frame.
We reshape $\boldsymbol{z}$ into $\boldsymbol{z_{full}} \in \mathbb{R}^{1 \times n_f*n_h*n_w \times d}$ as the input of the Global-Sequence Mamba block to capture spatial-first information. The Bidirectional-Mamba layer is used.

\textbf{Spatial and Temporal Mamba Blocks Interleaved.}
This particular variant leverages the Mamba module as a substitute for the traditional attention module within Transformer-based diffusion models for video generation, as noted in studies such as \cite{ma2024latte, guo2023animatediff, blattmann2023stable}. Illustrated in Fig. \ref{fig_architecture} (b), the backbone of this variant, known as Matten, is equipped with two types of Bidirectional-Mamba blocks: spatial Bidirectional-Mamba blocks and temporal Bidirectional-Mamba blocks. The spatial blocks are designed to solely capture spatial details among tokens that share identical temporal indices, whereas the temporal blocks are tasked with capturing information across different times within the same spatial coordinate. For effective spatial information processing, \(\boldsymbol{z}\) is restructured into \(\boldsymbol{z_s} \in \mathbb{R}^{n_f \times s \times d}\), which then serves as the input for the spatial Mamba block.

Then, we reshape $\boldsymbol{z_s}$ into $\boldsymbol{z_t} \in \mathbb{R}^{s \times n_f \times d}$ for the temporal Mamba block to process temporal information.

\textbf{Global-Sequence Mamba Block with Spatial-Temporal Attention Interleaved.}
Although Mamba demonstrates efficient performance in long-distance modeling, its advantages in shorter sequences modeling are not as pronounced~\cite{gu2023mamba}, compared to the attention operation in Transformer. 
Consequently, we have developed a hybrid block that leverages the strengths of both the attention mechanism and Mamba as illustrated in Fig. \ref{fig_architecture} (c), which integrates Mamba and Attention computations for both short and long-range modeling. Each block is composed of Spatial Attention computation, Temporal Attention computation, and a Global-Sequence Mamba scan in series. This design enables our model to effectively capture both the global and local information present in the latent space of videos.

\textbf{Global-Sequence Mamba Block with Temporal Attention Interleaved.} 

The scanning in the Global-Sequence Mamba block is continuous in the spatial domain but discontinuous in the temporal domain~\cite{li2024videomamba}. Thus, this variant has removed the Spatial Attention component, while retaining the Temporal Attention block. Consequently, by concentrating on a Spatial-First scan augmented with Temporal Attention shown in Fig. \ref{fig_architecture} (d), we strive to enhance our model's efficiency and precision in processing the dynamic facets of video data, thereby assuring robust performance in a diverse range of video processing tasks.

\begin{table*}[!t]
\centering
\resizebox{\linewidth}{!}{
\begin{tabular}{ccccccc}
\hline
Method          & Pretrained& FaceForensics & SkyTimelapse & UCF101 & Taichi-HD  & FLOPs (G)\\ \hline
MoCoGAN         & \XSolidBrush & 124.7         & 206.6        & 2886.9 & -   & -     \\
VideoGPT        & \XSolidBrush & 185.9         & 222.7        & 2880.6 & -  &-      \\
DIGAN           & \XSolidBrush & 62.5          & 83.11        & 1630.2 & 156.7 &-       \\
StyleGAN-V       &\XSolidBrush & 47.41         & 79.52        & 1431.0 & -   &-     \\
PVDM             &\XSolidBrush& 355.92        & 75.48        & 1141.9       & 540.2  & 1077      \\
MoStGAN-V         &\XSolidBrush& 39.70         & 65.30        & 1380.3      & -  &-\\
\hline
MoCoGAN-HD      &\CheckmarkBold  & 111.8         & 164.1        & 1729.6 & 128.1  & -  \\
LVDM          &\CheckmarkBold    & -             & 95.20        & 372.0       & \textbf{99.0} & 5718\\
Latte   & \CheckmarkBold& \textbf{34.00}         & 59.82        & 477.97 & 159.60 & 5572\\ \hline
Matten (ours)  & \XSolidBrush  & 45.01         & \textbf{53.56}        & \textbf{210.61} & 158.56 & 4008 \\
\hline
\end{tabular}}
\caption{FVD metrics for various video generation models across multiple datasets are presented. FVD scores for comparative baseline models, as reported in sources such as Latte, StyleGAN-V, or respective original publications, are included for reference. In this context, "Pretrained" refers to models that utilize a pretraining approach based on image generation techniques.}
\label{table_comparison_to_state-of-the-arts_fvd}
\end{table*}

\subsection{Conditional Way of Timestep or Class}
\label{sec:conditional_information_injection}

Drawing from the frameworks presented by Latte and DiS, we perform experiments on two distinct methodologies for embedding timestep or class information \(c\) into our model. The first method, inspired by DiS, involves treating \(c\) as tokens, a strategy we designate as \textit{conditional tokens}. The second method adopts a technique akin to adaptive normalization (AdaN) \cite{perez2018film, peebles2023scalable}, specifically tailored for integration within the Mamba block. This involves using MLP layer to compute parameters \(\gamma_c\) and \(\beta_c\) from \(c\), formulating the operation \(AdaN(f, c) = \gamma_c \text{Norm}(f) + \beta_c\), where \(f\) denotes the feature maps in the Mamba block.
Further, this adaptive normalization is implemented prior to residual connections of the Mamba block, implementing by the transformation \(RCs(f, c) = \alpha_c f + MambaScans(AdaN(f, c))\), with \(MambaScans\) representing the Bidirectional-Mamba scans within the block. We refer to this advanced technique as Mamba adaptive normalization (\textit{M-AdaN}), which seamlessly incorporates class or timestep information to enhance model responsiveness and contextual relevance.

\subsection{Analysis of Mamba and Attention}
\label{sec:compuation_anaylysis}

In summary, the hyperparameters of our proposed block encompass hidden size \(D\), expanded state dimension \(E\), and SSM dimension \(N\). All the settings of Matten are detailed in Table \ref{tab:model_params_flops}, covering different numbers of parameters and computation cost to thoroughly evaluate scalability performance. Specifically, the Gflop metric is analyzed during the generation of 16$\times$256$\times$256 unconditional videos, employing a patch size of \(p=2\). Consistent with \cite{gu2023mamba}, we standardize the SSM dimension \(N\) across all models at 16.

Both the SSM block within Matten and the self-attention mechanism in Transformer architectures are integral for effective context modeling. We provide a detailed theoretical analysis of computational efficiency as well. For a given sequence \(\textbf{X} \in \mathbb{R}^{1 \times J \times D}\) with the standard setting \(E=2\), the computational complexities of self-attention (SA), Feed-Forward Net (FFN) and SSM operations are calculated as follows:
\begin{align}
    \mathcal{O}(\text{SA}) &= 2J^2D,\\
    \mathcal{O}(\text{FFN}) &= 4JD^2,\\
    \mathcal{O}(\text{SSM}) &= 3J(2D)N + J(2D)N^2.
\end{align}
$3J(2D)N$ involves the calculation with $\overline{\mathbf{B}}$, $\mathbf{C}$, and $\mathbf{D}$, while $J(2D)N^2$ denotes the calculation with $\overline{\mathbf{A}}$. It demonstrates that self-attention's computational demand scales quadratically with the sequence length \(J\), whereas SSM operations scale linearly. Notably, with \(N\) typically fixed at 16, this linear scalability renders the Mamba architecture particularly apt for handling extensive sequences typical in scenarios like global relationship modeling in video data. When comparing the terms \(2J^2D\) and \(J(2D)N^2\), it is clear that the Mamba block is more computationally efficient than self-attention, particularly when the sequence length \(J\) significantly exceeds \(N^2\). For shorter sequences that focus on spatial and localized temporal relationships, the attention mechanism offers a more computationally efficient alternative when the computational overhead is manageable, as corroborated by empirical results.

\section{Experiments}

\begin{table}[t]
\centering
\resizebox{0.85\linewidth}{!}{
\begin{tabular}{ccccc}
\hline
           & Variant 1  & Variant 2  & Variant 3  & Variant 4  \\ \hline
Params (M) & 814  & 814  & 853  & 846 \\
FLOPs (G)  & 1590 & 1660 & 4008 & 3660 \\ \hline
\end{tabular}}
\caption{The parameter count and FLOPs (Floating-Point Operations) associated with various model variants of Matten.}
\label{tab:model_params_flops}
\end{table}

\begin{figure*}[t]
    \centering
    \includegraphics[width=\linewidth]{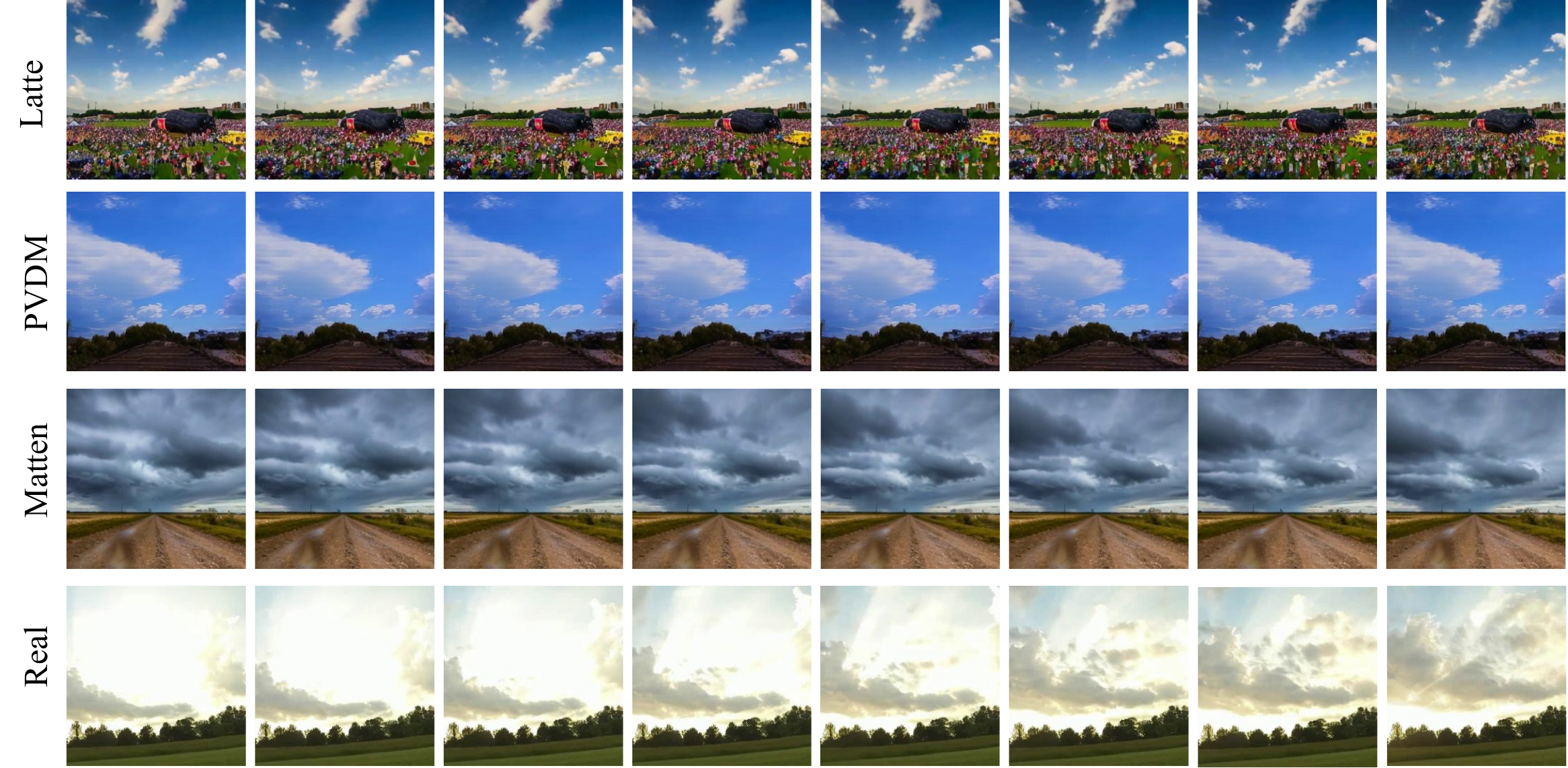}
    \caption{Sample videos from the different methods and real data on SkyTimelapse.} 
    \label{fig_sky}
\end{figure*}

\begin{figure*}[t]
    \centering
    \includegraphics[width=\linewidth]{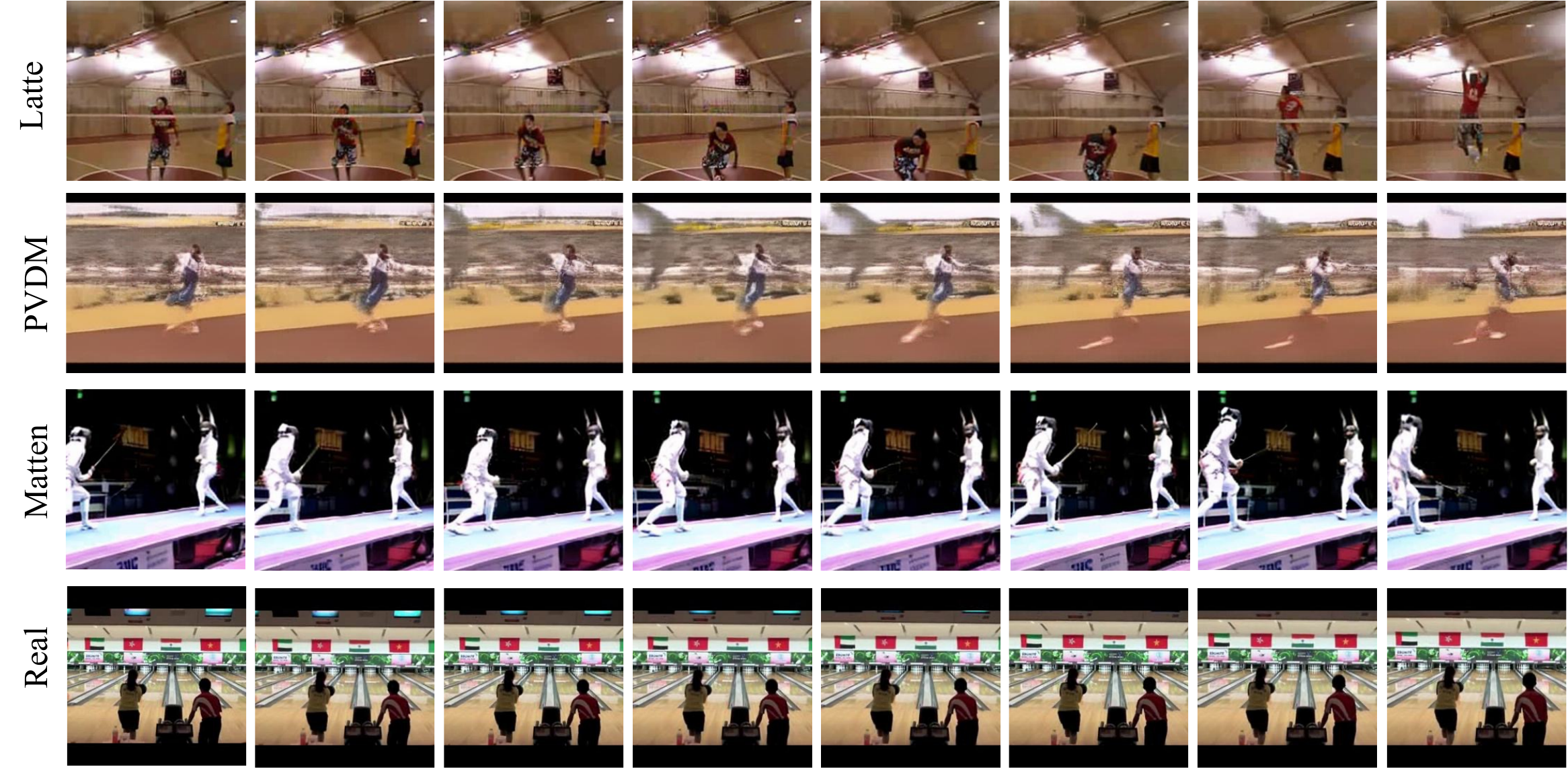}
    \caption{Sample videos generated using various methods on the UCF101 dataset, highlighting the visually appealing nature of the results. } 
    \label{fig_ucf}
\end{figure*}

\begin{figure*}[t]
    \centering
    \includegraphics[width=\linewidth]{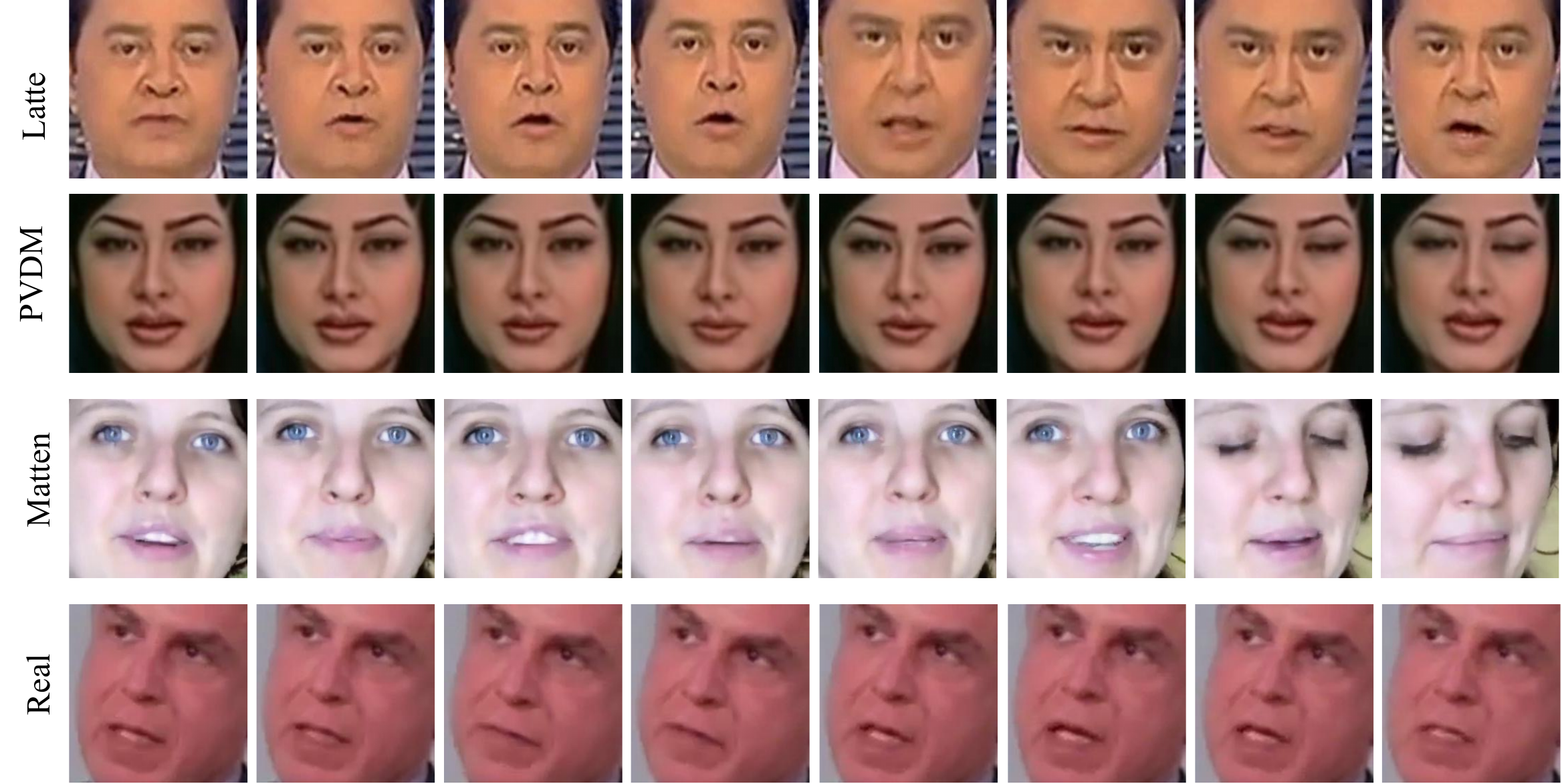}
    \caption{Sample videos generated using various methods on the FaceForensics dataset, highlighting the visually appealing nature of the results.} 
    \label{fig_ffs}
\end{figure*}

\begin{figure*}[t]
    \centering
    \includegraphics[width=\linewidth]{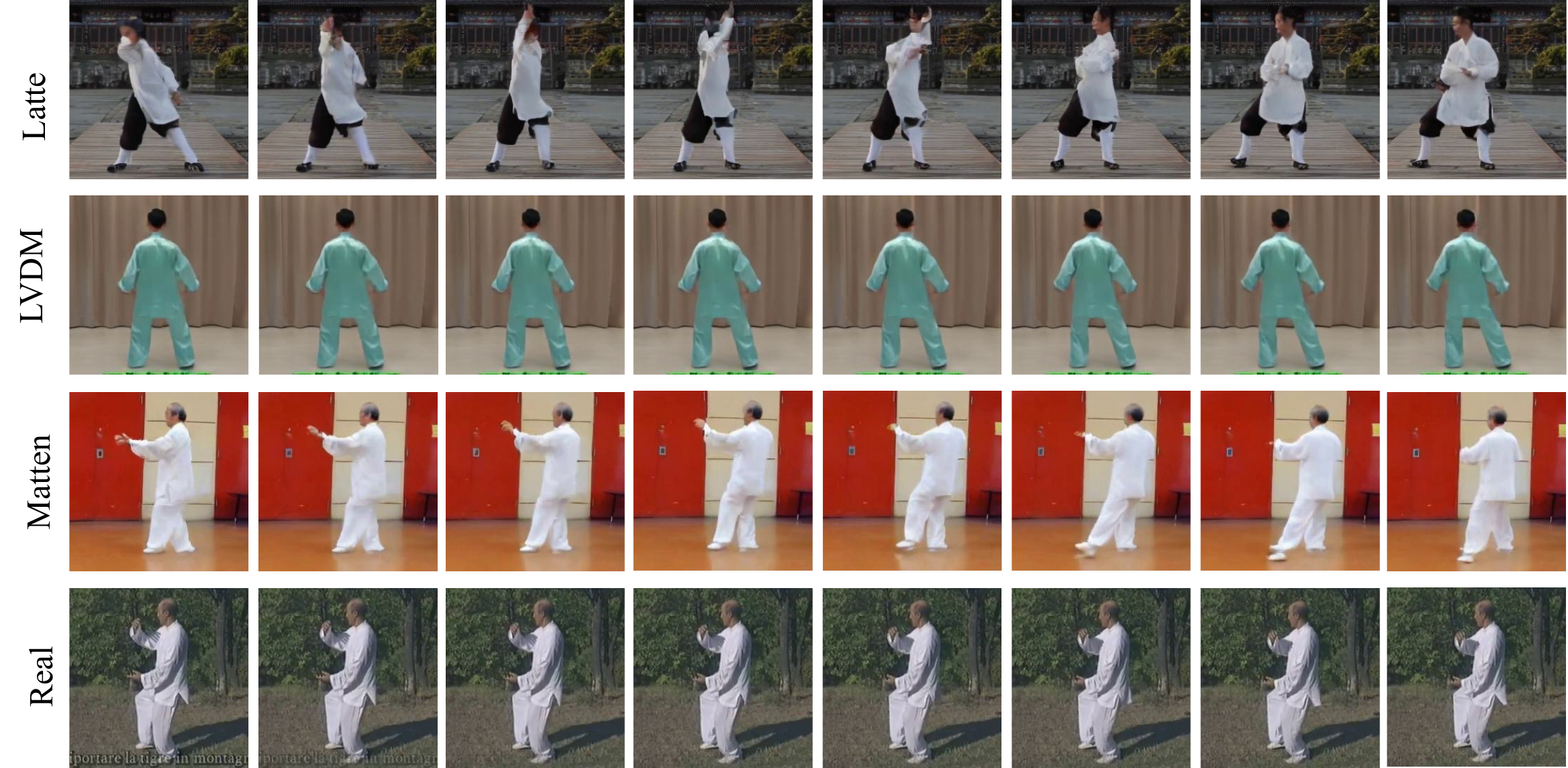}
    \caption{Sample videos generated using various methods on the Taichi-HD dataset, highlighting the visually appealing nature of the results.} 
    \label{fig_taichi}
\end{figure*}

This part first describes the experimental settings, including details about the datasets we used, evaluation metrics, compared methods, configurations of the Matten model, and specific implementation aspects. Following this, ablation studies are conducted to identify optimal practices and assess the impact of model size. The section concludes with a comparative analysis of our results on 4 common datasets against advanced video generation methods.

\begin{figure*}[t]
  \centering

  \subfloat[Model variants]{
      \label{fig_ablation:model_variants}\includegraphics[width=0.5\textwidth]{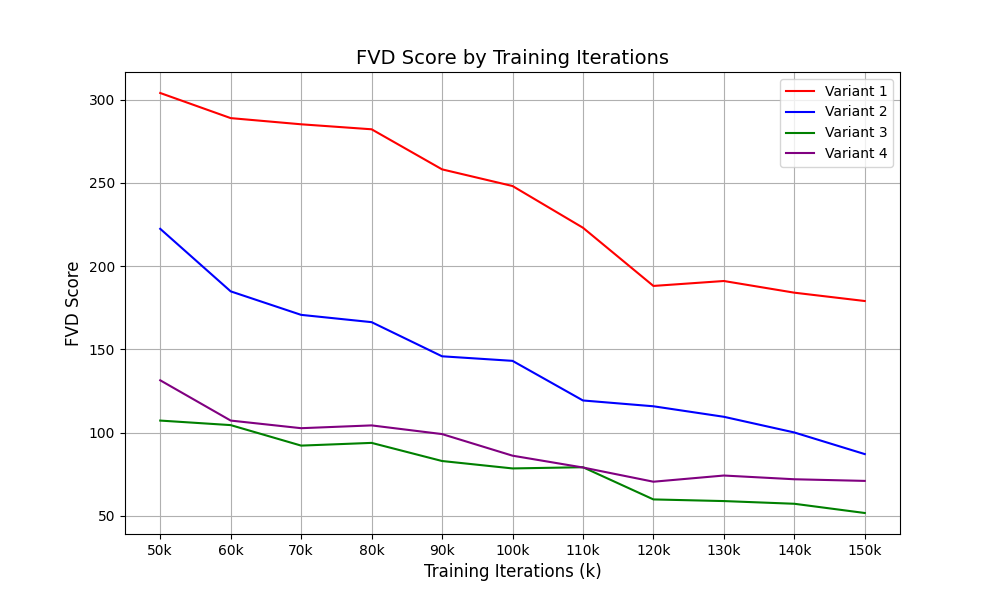}
      
  }
  \subfloat[Timestep-class conditional way]{
      \label{fig_ablation:conditional_information_injection}\includegraphics[width=0.5\textwidth]{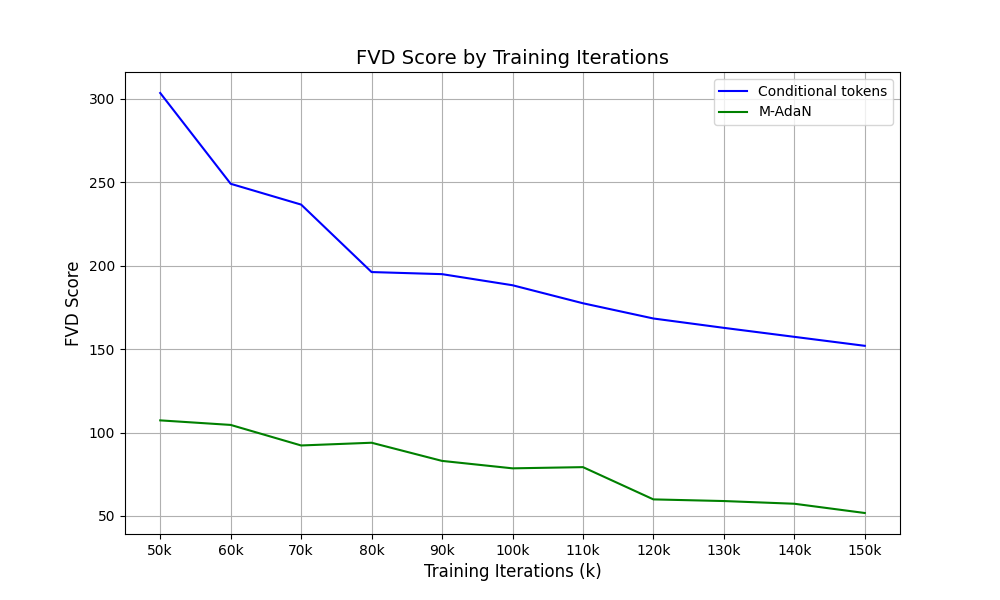}
  }

  \caption{Exploration of Design Choices Through Ablation Studies. We have conducted various ablation studies to identify optimal strategies for Mamba-based video diffusion models, focusing on improving FVD metrics on the SkyTimelapse dataset. For enhanced clarity, please magnify the displayed results.}
  \label{fig_ablation}
\end{figure*}
\begin{figure}[t]
\centering
\includegraphics[width=0.7\linewidth]{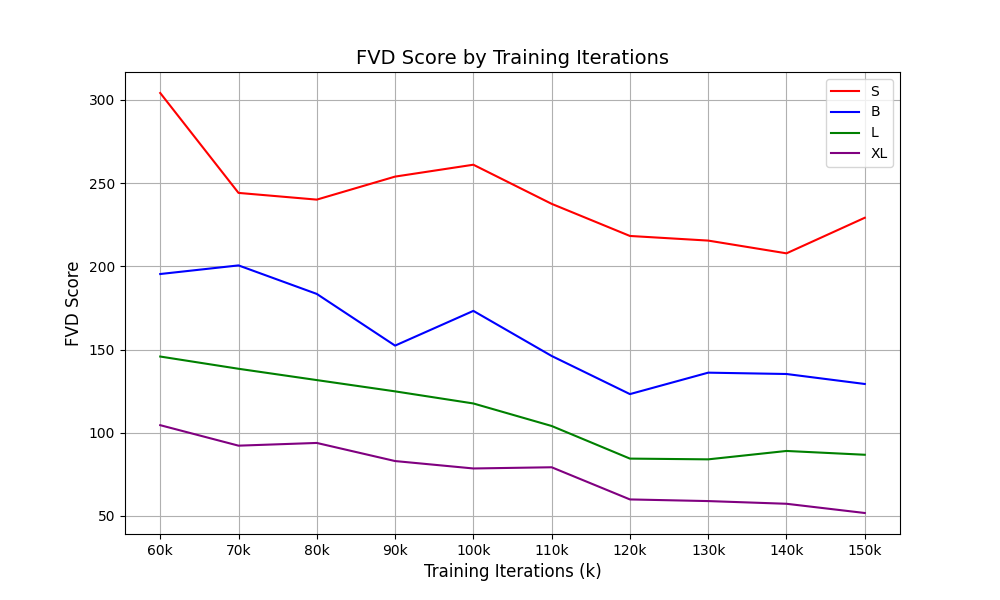}
\caption{The impact of varying model sizes on performance is notable. Generally, enlarging the model dimensions tends to markedly enhance its effectiveness.} 
\label{fig:model_size}
\end{figure}
\begin{table}[t]
\centering
\resizebox{0.85\linewidth}{!}{
\begin{tabular}{ccccc}
\hline
Model         & Layer numbers L & Hidden size D & SSM dimension N & Param    \\ \hline
Matten-S  & 12              & 384           & 16     & 35M  \\
Matten-B  & 12              & 768           & 16    & 164M \\
Matten-L  & 24              & 1024          & 16    & 579M \\
Matten-XL & 28              & 1152          & 16    & 853M \\ \hline
\end{tabular}}
\caption{Specifics of our model configurations adhere to the setups outlined for various model sizes following the ViT and DiT frameworks.}
\label{tab:model_size}
\end{table}

\subsection{Experimental Detail}
\textbf{Datasets Overview.} We engage in extensive experiments across four renowned and common datasets: FaceForensics \cite{rossler2018faceforensics}, SkyTimelapse \cite{xiong2018learning}, UCF101 \cite{soomro2012dataset}, and Taichi-HD \cite{siarohin2019first}. Following protocols established in Latte, we utilize predefined training and testing divisions. From these datasets, we extract video clips consisting of 16 frames, applying a sampling interval of 3, and resize each frame to a uniform resolution of 256x256 for our experiments.

\textbf{Evaluation Metrics.} For robust quantitative analysis, we adopt the Fréchet Video Distance (FVD) \cite{unterthiner2018towards}, recognized for its correlation with human perceptual evaluation. In compliance with the methodologies of StyleGAN-V, we determine FVD scores by examining 2,048 video clips, each containing 16 frames.

\textbf{Baseline Comparisons.} Our study includes comparisons with advanced methods to assess the performance of our approach quantitatively, including MoCoGAN \cite{tulyakov2018mocogan}, VideoGPT \cite{yan2021videogpt}, MoCoGAN-HD \cite{tian2021good}, DIGAN \cite{yu2022generating}, StyleGAN-V \cite{skorokhodov2022stylegan}, PVDM \cite{yu2023video}, MoStGAN-V \cite{shen2023mostgan}, LVDM \cite{he2023latent}, and Latte \cite{ma2024latte}. Unless explicitly stated otherwise, all presented values are obtained from the latest relevant studies: Latte, StyleGAN-V, PVDM, or the original paper.

\textbf{Matten Model Configurations.} Our Matten model is structured using a series of \(L\) Mamba blocks, with each block having a hidden dimension of \(D\). Inspired by the Vision Transformer (ViT) approach, we delineate four distinct configurations varying in parameter count, detailed in Table \ref{tab:model_size}.

\textbf{Implementation Specifics.} All ablation experiments adopt the AdamW optimizer, set at a fixed learning rate of \(1 \times 10^{-4}\). The sole augmentation technique applied is horizontal flipping. Consistent with prevailing strategies in generative modeling \cite{peebles2023scalable, bao2023all}, we employ the exponential moving average (EMA) of the model weights with a decay rate of 0.99 at the first 50k steps and the other 100k steps during the training process. The results reported are derived directly using the EMA-enhanced models. Additionally, the architecture benefits from the integration of a pre-trained variational autoencoder, sourced from Stable Diffusion v1-4.

\subsection{Ablation study}
\label{ablation_study}
In this part, we detail our experimental investigations using the SkyTimelapse dataset to assess the impact of various design modifications, model variations, and model sizes on performance, as previously introduced in Secs. \ref{sec:conditional_information_injection} and \ref{sec:the_details_of_Matten}.

\textbf{Timestep-Class Information Injection}
Illustrated in Fig. \ref{fig_ablation:conditional_information_injection}, the \textit{M-AdaN} approach markedly outperforms \textit{conditional tokens}. We surmise this difference stems from the method of integration of timestep or class information. \textit{Conditional tokens} are introduced directly into the model's input, potentially creating a spatial disconnect within the Mamba scans. In contrast, \textit{M-AdaN} embeds both timestep and class data more cohesively, ensuring uniform dissemination across all video tokens, and enhancing the overall synchronization within the model.

\textbf{Exploring Model Variants}
Our analysis of Matten’s model variants, as detailed in Sec. \ref{sec:the_details_of_Matten}, aims to maintain consistency in parameter counts to ensure equitable comparisons. Each variant is developed from the ground up. As depicted in Fig. \ref{fig_ablation:model_variants}, Variant 3 demonstrates superior performance with increasing iterations, indicating its robustness. Conversely, Variants 1 and 2, which focus primarily on local or global information, respectively, lag in performance, underscoring the necessity for a balanced approach in model design.

\textbf{Assessment of Model Size}
We experiment with four distinct sizes of the Matten model—XL, L, B, and S as listed in Tab. \ref{tab:model_size} on the SkyTimelapse dataset. The progression of their Fréchet Video Distances (FVDs) with training iterations is captured in Fig. \ref{fig:model_size}. There is a clear trend showing that larger models tend to deliver improved performance, echoing findings from other studies in image and video generation \cite{peebles2023scalable}, which highlight the benefits of scaling up model dimensions.

\subsection{Comparison Experiment}
\label{sec_comparison_to_state-of-the-art}
According to the findings from the ablation studies presented in Sec. \ref{ablation_study}, we have pinpointed the settings about how to design our Matten, notably highlighting the efficacy of model variant 3 equipped with \textit{M-AdaN}. Leveraging these established best practices, we proceed to conduct comparisons against contemporary state-of-the-art techniques.

\textbf{Qualitative Assessment of Results}
Figures \ref{fig_sky} through \ref{fig_taichi} display the outcomes of video synthesis using various methods across datasets such as UCF101, Taichi-HD, FaceForensics, and SkyTimelapse. Across these different contexts, our method consistently delivers realistic video generations at a high resolution of 256x256 pixels. Notable achievements include accurately capturing facial motions and effectively handling dynamic movements of athletes. Our model particularly excels in generating high-quality videos on the UCF101 dataset, an area where many other models frequently falter. This capability underscores our method's robustness in tackling complex video synthesis challenges.

\textbf{Quantitative results.} Tab. \ref{table_comparison_to_state-of-the-arts_fvd} presents the quantitative results of each comparative method. Overall, our method surpasses prior works and matches the performance of methods with image-pretrained weights, demonstrating our method's superiority in video generation.
Furthermore, our model attains roughly a 25\% reduction in flops compared to Latte, the latest Transformer-based model.
Given the abundance of released pre-trained U-Net-based (Stable Diffusion, SDXL) or Transformer-based (DiT, PixArt) image generation models, these U-Net-based or Transformer-based video generation models can leverage these pre-trained models for training. However, there are no released, pre-trained Mamba-based image generation models yet, so our model has to be trained from scratch. We believe that once Mamba-based image generation models become available, they will be of great help in training our Matten.

\section{Conclusion}

This paper proposes a simple diffusion method for video generation, Matten, with the Mamba-Attention structure as the backbone for generating videos. To explore the quality of Mamba for generating videos, we explore different configurations of the model, including four model variants, time step and category information injection, and model size. Extensive experiments demonstrate that Matten excels in four standard video generation benchmarks and displays impressive scalability.

{\small
    \bibliographystyle{unsrt}
\bibliography{neurips_2023}
  }
\end{document}